\documentclass[sigconf,natbib=true]{acmart}
\AtBeginDocument{%
  }

\copyrightyear{2026}
\acmYear{2026}
\setcopyright{cc}
\setcctype{by-nc-nd}
\acmConference[SIGIR '26]{Proceedings of the 49th International ACM SIGIR Conference on Research and Development in Information Retrieval}{July 20--24, 2026}{Melbourne, VIC, Australia}
\acmBooktitle{Proceedings of the 49th International ACM SIGIR Conference on Research and Development in Information Retrieval (SIGIR '26), July 20--24, 2026, Melbourne, VIC, Australia}
\acmDOI{10.1145/3805712.3808368}
\acmISBN{979-8-4007-2599-9/2026/07}

\usepackage{enumitem} % Make sure this is in your preamble
\usepackage{float}
\usepackage{stfloats}
\usepackage{latexsym}
\usepackage{listings}
\usepackage{longtable}
\usepackage{multicol}
\usepackage{url,tcolorbox}
\usepackage{graphicx}
\usepackage{wrapfig}
\usepackage{natbib}
\usepackage{booktabs,fvextra}
\usepackage{comment}
\usepackage{subcaption}
\usepackage{verbatim}
\usepackage{longtable}
\usepackage{tikz}
\usepackage{array}
\usepackage{enumitem,verbatimbox}
\usepackage{xspace}
\usepackage{algorithm} 
\usepackage{algpseudocode}
\usepackage{multirow}
\usepackage[normalem]{ulem}

\lstdefinelanguage{json-like}{
  basicstyle=\ttfamily,
  numbers=left,
  numberstyle=\tiny\color{gray},
  numbersep=5pt,
  showstringspaces=false,
  breaklines=true,
  frame=single,
  morestring=[b]",
  stringstyle=\color{blue},
  morecomment=[l]{//},
  commentstyle=\color{green},
  morecomment=[s]{/*}{*/},
  morekeywords={content, role}
}

\newcommand{\tool}{\textsc{Peerispect}}

\begin{document}

%%
%% The "title" command has an optional parameter,
%% allowing the author to define a "short title" to be used in page headers.
\title{\tool: Claim Verification in Scientific Peer Reviews}

%%
%% The "author" command and its associated commands are used to define
%% the authors and their affiliations.
%% Of note is the shared affiliation of the first two authors, and the
%% "authornote" and "authornotemark" commands
%% used to denote shared contribution to the research.

\author{Ali Ghorbanpour}
\orcid{0009-0006-7383-8105}
\affiliation{%
  \institution{Reviewerly}
    \city{Toronto}
    \state{Ontario}
    \country{Canada}
  }
  
% \email{aligh@reviewer.ly}
\author{Soroush Sadeghian}
\orcid{0009-0005-2172-7617}
\affiliation{%
  \institution{Reviewerly}
    \city{Toronto}
    \state{Ontario}
    \country{Canada}
  }
% \email{soroushsa@reviewer.ly}

\author{Alireza Daghighfarsoodeh}
\orcid{}
\affiliation{%
  \institution{Reviewerly}
    \city{Toronto}
    \state{Ontario}
    \country{Canada}
  }
% \email{sajad@reviewer.ly}

 \author{Sajad Ebrahimi}
\authornote{Corresponding author. Email: \texttt{s.ebrahimi@utoronto.ca}}
 \orcid{0009-0003-1630-3938}
 \affiliation{%
   \institution{University of Toronto, Reviewerly}
 \city{Toronto}
 \state{Ontario}
  \country{Canada}
   }
% \email{sajad@reviewer.ly}

\author{Negar Arabzadeh}
\orcid{0000-0002-4411-7089}
\affiliation{%
  \institution{Reviewerly, UC Berkeley}
    \city{Berkeley}
    \state{California}
    \country{United States}
  }
% \email{negara@reviewer.ly}

\author{Seyed Mohammad Hosseini}
\affiliation{%
  \institution{Reviewerly}
    \city{Toronto}
    \state{Ontario}
    \country{Canada}
  }
% \email{negara@reviewer.ly}

\author{Ebrahim Bagheri}
\orcid{0000-0002-5148-6237}
\affiliation{%
  \institution{University of Toronto, Reviewerly}
    \city{Toronto}
    \state{Ontario}
    \country{Canada}
  }
% \email{ebrahim.bagheri@utoronto.ca}

%%
%% By default, the full list of authors will be used in the page
%% headers. Often, this list is too long, and will overlap
%% other information printed in the page headers. This command allows
%% the author to define a more concise list
%% of authors' names for this purpose.
\renewcommand{\shortauthors}{Ali Ghorbanpour et al.}

\begin{abstract}
Peer review is central to scientific publishing, yet reviewers frequently include claims that are subjective, rhetorical, or misaligned with the submitted work. Assessing whether review statements are factual and verifiable is crucial for fairness and accountability. At the scale of modern conferences and journals, manually inspecting the grounding of such claims is infeasible. We present \textbf{\tool}, an interactive system that operationalizes claim-level verification in peer reviews by extracting check-worthy claims from peer reviews, retrieving relevant evidence from the manuscript, and verifying the claims through natural language inference. Results are presented through a visual interface that highlights evidence directly in the paper, enabling rapid inspection and interpretation. \tool\ is designed as a modular Information Retrieval (IR) pipeline, supporting alternative retrievers, rerankers, and verifiers, and is intended for use by reviewers, authors, and program committees. We demonstrate \tool\ through a live, publicly available demo\footnote{\url{https://app.reviewer.ly/app/peerispect}} and API services\footnote{\url{https://github.com/Reviewerly-Inc/Peerispect}}, accompanied by a video tutorial\footnote{\url{https://www.youtube.com/watch?v=pc9RkvkUh14}}.
\end{abstract}

%%
%% The code below is generated by the tool at http://dl.acm.org/ccs.cfm.
%% Please copy and paste the code instead of the example below.
%%

% \ccsdesc[500]{Computing methodologies~Natural language processing}

% %%
% %% Keywords. The author(s) should pick words that accurately describe
% %% the work being presented. Separate the keywords with commas.

%%
%% This command processes the author and affiliation and title
%% information and builds the first part of the formatted document.
\maketitle

\section{Introduction}

Peer review remains the cornerstone of scholarly publishing, ensuring that only research meeting standards of novelty, rigor, and clarity enters the scientific record. Despite its importance, peer review is far from flawless \cite{smith2006peer, tennant2017multi, drozdz2024peer, arabzadeh2025building}. Reviewers frequently include claims that go beyond objective evaluation, sometimes misinterpreting the content of the submission, overstating limitations, or questioning results in ways not fully supported by the manuscript itself \cite{lee2013bias, bell2024scholarly}. Such issues can compromise the fairness of reviews, mislead editors and program committees, and ultimately slow down the progress of scientific research \cite{garg2020problems}. While prior work on factuality assessment and claim verification has largely focused on pipelines for general fact-checking or scientific fact verification \cite{thorne2018fever,scire2024fenice,worledge2024extractive}, there has been little effort to address the unique setting of verifying reviewer statements against a manuscript.

A substantial body of work has documented limitations of the peer review process, including bias, inconsistency, and noise \cite{smith2006peer, tennant2017multi, drozdz2024peer, lee2013bias, bell2024scholarly, ebrahimi2025exharmony, arabzadeh2024reviewerly}. Reviewers may misinterpret experimental settings, overlook details, or overstate shortcomings, leading to claims that are not fully grounded in the paper itself \cite{wadden2020scifact, dasigi2021datasetinformationseekingquestionsanswers, arabzadeh2026can}. These issues are particularly consequential because review statements directly influence editorial decisions, acceptance outcomes, and author revisions. Yet, prior works on peer-review analysis have largely focused on systemic properties of the review process and high-level assessments of review quality. Studies such as \citet{lee2013bias} analyze sources of bias, including reviewer identity and institutional affiliation, while \citet{tennant2017multi} provide a broad examination of transparency and fairness. Even recent benchmarks like RottenReviews \cite{ebrahimi2025rottenreviews} or models for review constructiveness \cite{sadallah2025good} primarily operate at the document or discourse level. They assess overall tone or utility but do not perform the grounded, claim-level verification necessary to identify the specific inconsistencies described above.
Addressing this verification gap presents a unique technical challenge. While the information retrieval community has developed mature techniques for factuality assessment, claim verification, and evidence retrieval, most prior work has focused on open domain fact checking or scientific claim verification against large external corpora \cite{thorne2018fever, scire2024fenice, worledge2024extractive}. In contrast, verifying reviewer claims requires grounding statements against a single submitted document, giving way to a distinct and underexplored retrieval problem.

\begin{figure}[t]
    \centering
    \includegraphics[width=.9\linewidth]{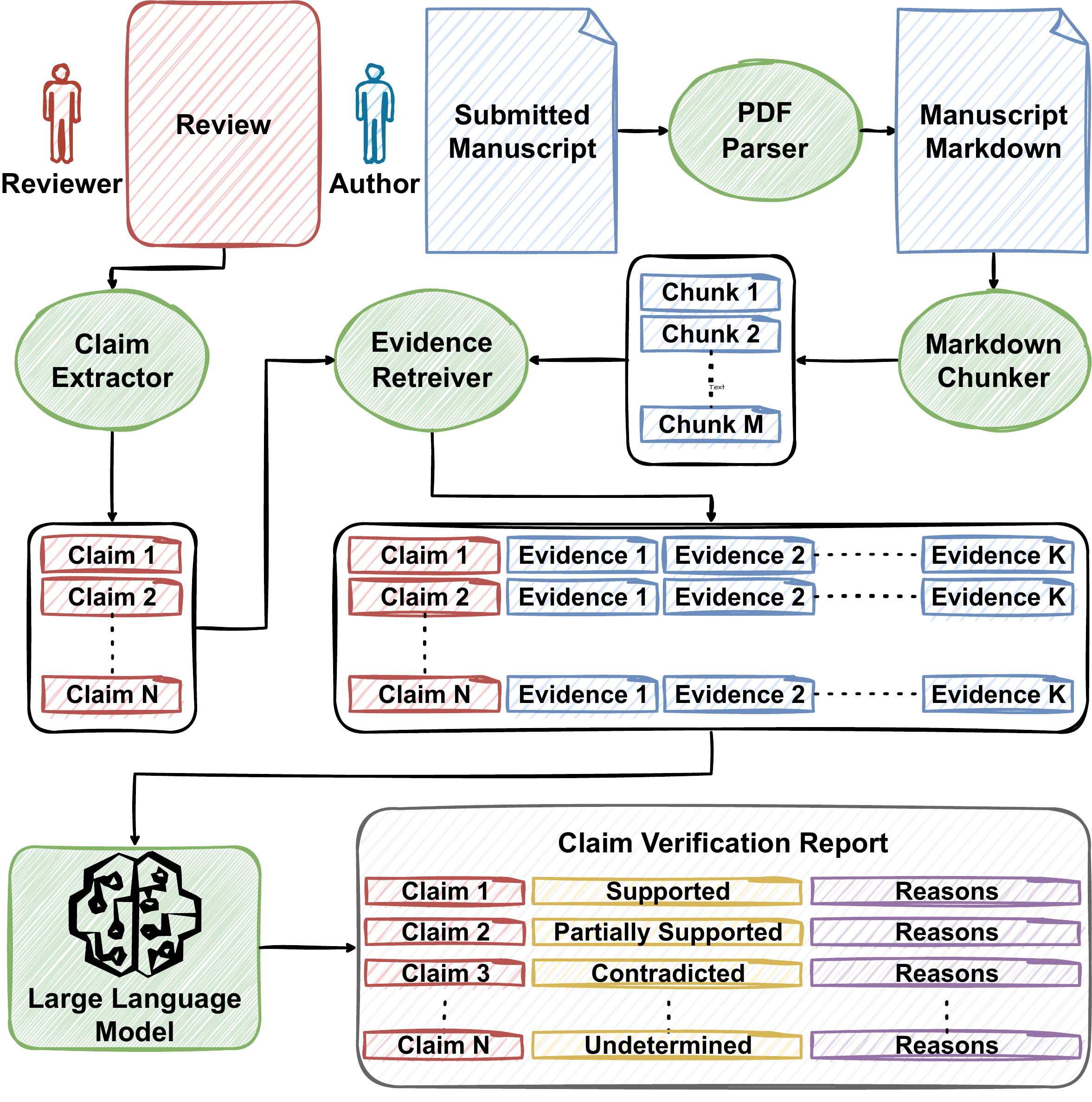}
    \caption{\tool\ architecture and processing pipeline. The system consists of four stages: data acquisition, claim extraction, evidence retrieval, and claim verification.}
    \vspace{-2em}
    \label{fig:architecture}
\end{figure}

The need for scalable support has intensified as the volume of scientific submissions continues to grow \cite{leopold2015increased, azad2024publication}. Major conferences now receive tens of thousands of papers per cycle. For example, AAAI received approximately 12,000 submissions in 2025, which increased to roughly 23,000 submissions in 2026 \cite{aaai2026reviewprocessupdate}. As submission volumes increase, so does the volume of review text, making it infeasible for program committees and editors to manually inspect whether review claims are well grounded. From an information retrieval perspective, this setting naturally raises the question of whether reviewer claims can be treated as information seeking queries over the submitted paper, and whether retrieval and inference techniques can be used to assess their grounding automatically.

In this work, we introduce \tool, an interactive system that operationalizes claim level verification in peer reviews by framing reviewer statements as queries over the manuscript. The system extracts check worthy claims from reviews, retrieves the most relevant passages from the paper, and verifies alignment via natural language inference. The output indicates whether a claim is supported, partially supported, contradicted, or undetermined, and presents these judgments through a visual interface that highlights evidence directly in the manuscript.

% \tool\ is designed as a modular and extensible information retrieval pipeline rather than a fixed tool. Each stage of the system, including claim extraction, retrieval, reranking, and verification, can be replaced or extended independently. This design enables researchers and practitioners to experiment with alternative retrievers, ranking strategies, or inference models while preserving the overall workflow. 

The system supports multiple stakeholders in the peer review process. Reviewers can inspect whether their factual statements are grounded in the manuscript. Authors can identify which review claims are supported or contradicted when preparing rebuttals or revisions. Program committees and editors can use the system as a discretionary aid to maintain review quality at scale. By making claim evidence alignments explicit, \tool\ increases transparency while preserving human judgment as the final authority.

Beyond live demonstration, we empirically validate the underlying pipeline using two complementary datasets that reflect both controlled and real world conditions. The \texttt{Controlled Manuscript Claims (CMC)} benchmark consists of 500 paper derived claims extracted from 50 ICLR 2024 manuscripts, representing an upper bound scenario in which all claims are supported by construction. The \texttt{Real World Review Claims (RRC)} comprises 150 manually annotated reviewer claims from 25 papers. comprising 150 manually annotated reviewer claims from 25 papers. These complementary datasets allow us to rigorously assess evidence retrieval and verification accuracy, ensuring the demo reflects a tested, reliable system rather than a purely illustrative prototype.

Our \tool\ tool demonstrates how retrieval, ranking, and inference techniques can be integrated to support accountability and transparency in peer review workflows.
By treating review verification as a document grounded information retrieval problem and providing an interactive and extensible system, this work positions peer review analysis as a practical and impactful application area for the IR research community.

\section{Tool Description}
\label{sec:system}

\begin{figure}[t]
    \centering
    \vspace{-1em}
    \includegraphics[width=0.5\textwidth]{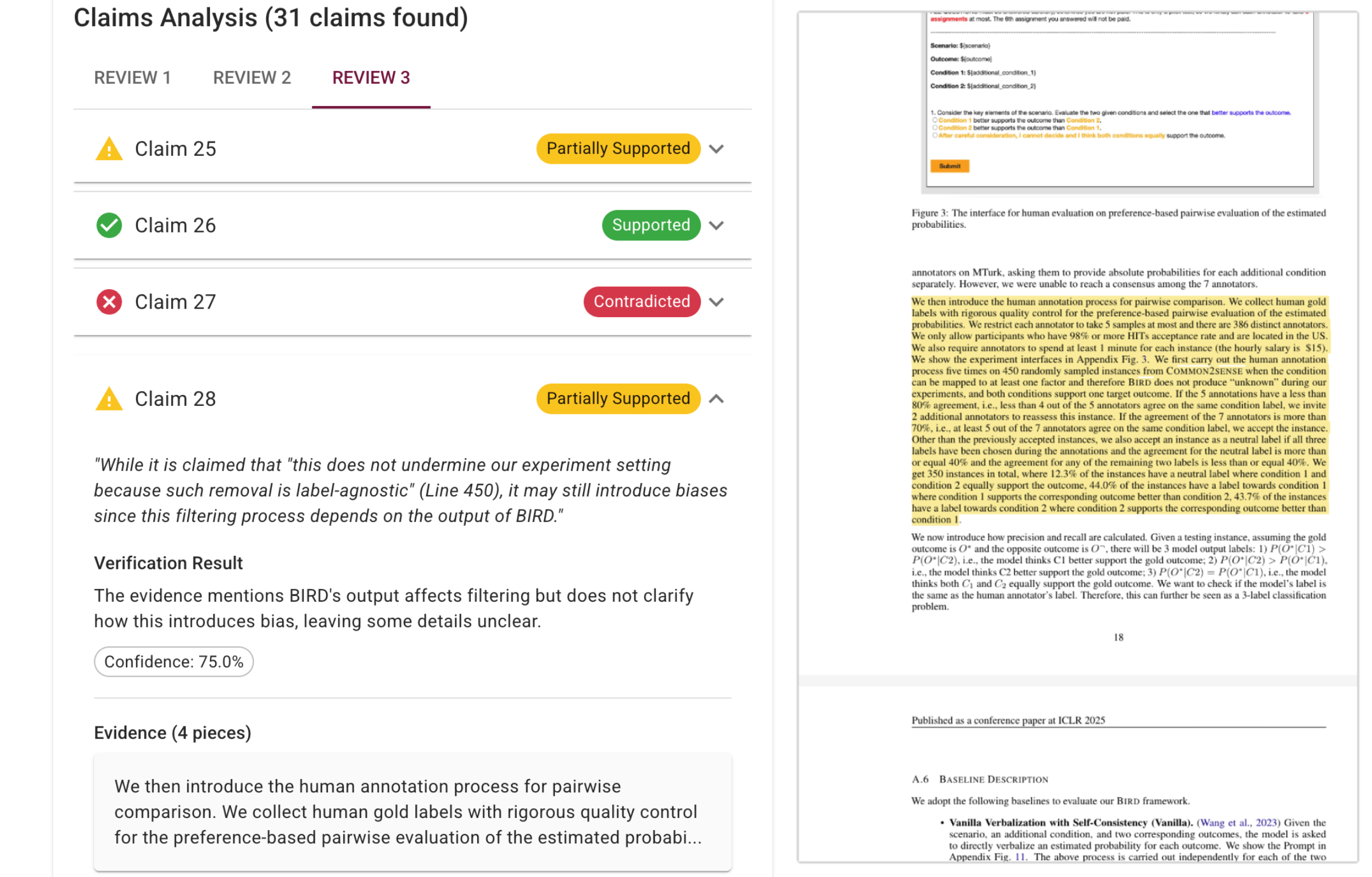}
    \caption{Screenshot of the \tool\ interface.}
    \vspace{-1em}
    \label{fig:screenshot}
\end{figure}

\tool\ is an interactive system that operationalizes claim level verification in peer reviews through a modular information retrieval pipeline. The system processes a manuscript together with its associated reviews and produces structured annotations that characterize the grounding of individual review claims with respect to the paper. The design emphasizes modularity and extensibility, allowing alternative retrieval, ranking, and verification components to be integrated without altering the overall workflow. Figure~\ref{fig:architecture} presents an overview of the pipeline, while Figure~\ref{fig:screenshot} illustrates the user facing interface. The pipeline consists of four conceptual stages: \textit{data ingestion, claim extraction, evidence retrieval}, and\textit{ claim verification}. Each stage corresponds to a well defined IR or NLP subtask and can be independently modified or replaced. This design reflects the goal of presenting \tool\ as a reusable research artifact rather than a fixed end to end model.

\textbf{Data ingestion} The \textit{data ingestion} stage prepares the manuscript and reviews for downstream processing. Manuscripts are segmented into semantically coherent textual units that serve as retrievable documents, while reviews are normalized into sentence level inputs. Although \tool\ supports multiple ingestion modes, including direct document upload and integration with external review platforms such as OpenReview.net, these mechanisms are treated as interchangeable front ends whose primary purpose is to supply clean textual representations to the retrieval pipeline.

\textbf{Claim extraction.} The \textit{claim extraction} stage identifies check worthy factual claims from review text. Peer reviews typically interleave factual observations with subjective judgments and prescriptive feedback. To isolate verifiable content, \tool\ decomposes review sentences into candidate spans and filters them using a classifier trained to distinguish factual, evidence seeking statements from opinionated or rhetorical language. This process follows prior observations that early removal of subjective content improves downstream factuality assessment \cite{scire2024fenice}. The remaining spans are normalized into atomic claims by resolving coreference and removing hedging expressions, yielding a set of minimal claims suitable for retrieval based verification.

\textbf{Evidence retrieval.} In the \textit{evidence retrieval} stage, each extracted claim is treated as a query over the manuscript. The paper is indexed as a collection of fixed length passages, and relevant passages are retrieved using a combination of semantic similarity and reranking. In the deployed demo, dense retrieval is performed using \texttt{MiniLM-L6-v2} embeddings \cite{reimers-2019-sentence-bert} with FAISS based nearest neighbor search \cite{douze2024faiss}. Retrieved candidates are then reranked using a cross encoder that jointly models the claim and passage \cite{nogueira2020passagererankingbert}. While this configuration is used in the live system, the retrieval stage is explicitly designed to support alternative sparse, dense, or hybrid retrieval strategies.

\textbf{Claim Verification.} The final stage performs claim verification by assessing the relationship between each claim and its retrieved evidence. Following standard formulations in claim verification and natural language inference \cite{thorne2018fever, worledge2024extractive}, the claim is treated as a hypothesis and the evidence passages as premises. An LLM based verifier assigns one of four labels indicating support, partial support, contradiction, or uncertainty. These labels provide a structured characterization of claim grounding that can be inspected by users and consumed by downstream analysis.

\textbf{\tool} \textbf{output.} The output of \tool\ is presented through an interactive interface that aligns claims, evidence, and verification outcomes. Claims are listed alongside their labels, and corresponding evidence passages are highlighted directly within the manuscript view. This design enables users to quickly inspect retrieval and verification behavior, making the system suitable for practical use. \tool\ exposes claim verification in peer review as a document grounded information retrieval problem and provides an extensible platform for studying this task in realistic review settings.

\begin{table}[t!]
\centering
\footnotesize
\caption{
Dataset statistics for the two evaluation benchmarks \texttt{CMC} and \texttt{RRC}. 
}
\label{tab:dataset_stats}
\resizebox{\columnwidth}{!}{
\begin{tabular}{@{}lccc@{}}
\toprule
\textbf{Dataset} & \textbf{\#Papers} & \textbf{\#Claims} & \textbf{Description} \\ 
\midrule
CMC & 50  & 500  & Claims extracted from manuscripts \\ 
% B2: Reviewer-Derived & 25  & 800  & Reviews + author responses \\ 
RRC   & 25  & 150  & Manually labeled from author Reviewer Interactions\\ 
\bottomrule
\end{tabular}
}
\vspace{-2em}
\end{table}

\section{Implementation Details}
\label{sec:impl}

\tool\ boasts a web interface that sits over its service-oriented architecture. The design emphasizes modularity, scalability to support interactive use in real peer review scenarios. 
The backend, implemented in FastAPI, orchestrates data flow across modules, exposes a REST API, and manages communication with external services such as OpenReview.net. The frontend, built in React with react-pdf-highlighter, renders PDF documents and highlights evidence passages for each claim returned by the API. We integrate the VLLM inference engine \cite{kwon2023efficient}, which enables us to easily set up local LLM installations that are both efficient and scalable, while maintaining data privacy by avoiding reliance on external APIs. For the live demo, we use \texttt{Qwen-2.5-7B} \cite{yang2025qwen3}, but the API supports integration with any model available through VLLM. To ensure portability and reproducibility, \tool\ is packaged using Docker, with each service (backend, frontend, and model server) running in its own container to simplify deployment and resource management. This containerized setup allows \tool\ to be deployed on local servers for development or scaled to cloud environments for broader accessibility.

\section{Empirical Validation}
\label{sec:evaluation}

For \tool\ to be useful in real peer-review workflows, its behavior must be predictable and trustworthy. Rather than aiming for a comprehensive research evaluation, our goal in this section is to provide practical evidence that the system behaves reliably under conditions that resemble actual conference and journal reviewing. To this end, we built two complementary benchmarks from real ICLR 2024 submissions and their OpenReview discussions and used them to assess \tool.

\subsection{Benchmarks Derived from Real Peer Review}
\label{benchmarks}

All data is drawn from publicly available OpenReview entries for ICLR~2024. We sampled 50 submissions, stratified across accepted (oral), accepted (poster), and rejected papers, and collected their PDFs, metadata, and full review-rebuttal threads. Table~\ref{tab:dataset_stats} summarizes the resulting corpus.

\subsubsection{Controlled Manuscript Claims Benchmark (\texttt{CMC})}

To understand how \tool\ behaves in an idealized setting where every claim is guaranteed to be grounded in the manuscript, we constructed the Controlled Manuscript Claims benchmark (\texttt{CMC}). Using \texttt{OpenAI-o4-mini}, we decomposed each manuscript into atomic factual units and randomly sampled 10 claims per paper, yielding 500 instances in total. Because each claim is taken directly from the manuscript, every instance receives a gold label of \textit{Supported}.

This controlled setup serves two purposes. First, it allows us to check whether the retrieval component reliably surfaces the paragraphs that originally expressed each claim. Second, it lets us observe how the verifier behaves when it receives either oracle passages or system-retrieved passages as evidence. In practice, \texttt{CMC} acts as an upper-bound, low-ambiguity scenario where any errors can be attributed to retrieval or verification behavior rather than to unclear ground truth.

\subsubsection{Real-World Review Claims Benchmark (\texttt{RRC})}

Real peer reviews are less tidy. Reviewer comments may mix interpretations, paraphrases, and partially grounded critiques, and they often get clarified only through back-and-forth discussion with the authors. To capture this setting, we built the Real-World Review Claims benchmark (\texttt{RRC}) by randomly selecting 25 papers from the same 50 OpenReview submissions.
We applied \texttt{OpenAI-o4-mini} to convert reviewer comments into atomic claims, and appended the author response thread as context. From which, we sampled 150 claim-dialog pairs and manually assigned one of the following four labels:

\begin{itemize}[leftmargin=1.2em]
  \item \textbf{Supported:} when the author explicitly agrees with the reviewer or cites the manuscript in a way that confirms the claim.
  \item \textbf{Contradicted:} when the author disputes the claim and refers to manuscript-grounded evidence.
  \item \textbf{Partially Supported:} when the author qualifies the claim (e.g., “only under condition X” or “only for subset Y”).
  \item \textbf{Undetermined:} when the dialog does not resolve the issue or centers on subjective or policy-level statements.
\end{itemize}

\noindent Here, both the formulation of the claim and the relevance of evidence are less clear-cut. Thus we evaluate the full pipeline and examine whether \tool\ assigns labels that align with the human annotations. \texttt{RRC} is meant to reflect the conditions under which a reviewer, author, or editor would actually use the tool.

\subsection{Evaluating Different Configurations}

We use the two benchmarks in complementary ways. \texttt{CMC} is used to probe individual components, while \texttt{RRC} is used to understand end-to-end behavior under noisy, discourse-driven conditions.
On \texttt{CMC}, we vary retrieval strategies and evidence selection to see how design choices affect the tool’s ability to recover the original manuscript passages and assign the expected \textit{Supported} label. On \texttt{RRC}, we focus on the configurations that remain robust when claims are paraphrased, partially grounded, or indirectly linked to the text.

\textbf{Retrievers.} We compare three practical retrieval configurations that we considered for deployment in the demo: a sparse retriever (BM25), a dense retriever, and a sparse retriever followed by a cross-encoder reranker. We also examine the behavior of the cross-encoder when applied directly to the top-20 candidates from the sparse + reranker setup.

\textbf{LLMs.} For verification, we experimented with \texttt{Qwen-2.5-3B} and \texttt{Qwen-2.5-7B} within an otherwise identical pipeline to understand how model size affects stability. For all configurations, the verifier receives the top-3 passages per claim as evidence. We report accuracy and recall on both oracle and system-retrieved evidence (Table~\ref{tab:combined}), not as a leaderboard, but to give a concrete sense of how dependable the system is under different design choices. The configuration used in the live demo is chosen based on this analysis.

 \section{Findings and Observations}
\label{sec:results}

\begin{table}[t!]
\centering
\small
\caption{
Performance comparison across \texttt{CMC} and \texttt{RRC} benchmarks with different retrievers and LLMs.
}
\vspace{-1em}
\label{tab:combined}

\resizebox{0.5\textwidth}{!}{
\begin{tabular}{@{}llcccccc@{}}
\toprule
\multirow{2}{*}{\textbf{Model}} & 
\multirow{2}{*}{\textbf{Retriever}} &
\multicolumn{3}{c}{\textbf{CMC}} &
\multicolumn{3}{c}{\textbf{RRC}} \\
\cmidrule(lr){3-5} \cmidrule(lr){6-8}
 &  & ACC & Recall &  & ACC & Recall &  \\ 
\midrule

\multirow{5}{*}{Qwen-2.5-3b}
    & BM25                         & 0.728 & \textbf{0.486} &  & 0.220 & -- &  \\
    % & sbert naive                   & 0.614 & 0.418 &  & 0.193 & -- &  \\
    & Dense Retriever                         & 0.618 & 0.418 &  & 0.193 & -- &  \\
    & Dense + Reranker        & 0.740 & 0.476 &  & \textbf{0.247} & -- &  \\
    & Sparse + Reranker    & \textbf{0.744} & 0.478 &  & \textbf{0.247} & -- &  \\

\midrule
\multirow{5}{*}{Qwen-2.5-7b}
    & BM25                         & \textbf{0.905 }& \textbf{0.486} &  & 0.247 & -- &  \\
    % & sbert naive                   & 0.806 & 0.418 &  & 0.247 & -- &  \\
    & Dense Retriever                         & 0.804 & 0.418 &  & 0.247 & -- &  \\
    & Dense + Reranker        & 0.880 & 0.469 &  & \textbf{0.287} & -- &  \\
    & Sparse + Reranker    & 0.896 & 0.478 &  & 0.260 & -- &  \\

\bottomrule
\end{tabular}
}
\vspace{-1em}

\end{table}

The two benchmarks introduced in Section~\ref{benchmarks} allow us to observe how \tool\ behaves in settings that tool users are likely to encounter in practice. \texttt{CMC} represents a controlled, low-ambiguity regime where all claims are supported by construction. \texttt{RRC} captures the more challenging reality of noisy reviewer language and partially grounded comments. Below we summarize the patterns that matter most for users of the demo.

\noindent \textit{Observations on \texttt{CMC.}} Because every claim in \texttt{CMC} originates from the manuscript itself, a well-behaved system should retrieve the corresponding passage and mark the claim as supported. Consistent with this expectation, accuracy on \texttt{CMC} is substantially higher than on \texttt{RRC} across all configurations (Table~\ref{tab:combined}). We see two clear trends:

\begin{itemize}[leftmargin=1.25em]
\item Larger verification models (\texttt{Qwen-2.5-7B}) yield noticeably higher accuracy than smaller ones (\texttt{Qwen-2.5-3B}), suggesting that additional capacity helps with nuanced NLI judgments once relevant evidence is available.
\item Sparse retrieval with BM25 performs very strongly in this setting. Since claims are extracted from the manuscript, their wording tends to be lexically close to the source passages, and BM25 achieves high recall and accuracy.
\end{itemize}

Because \texttt{CMC} also includes an oracle-evidence condition, it acts as a sanity check. Any remaining errors point to either retrieval misses or verifier misclassifications rather than to ambiguous labels. This gives users confidence that, when claims are directly tied to the text, \tool\ behaves predictably.

\noindent \textit{Observations on \texttt{RRC}.} \texttt{RRC} is deliberately closer to how reviewers actually write. Claims can be vague, compressed, or only partially grounded in the manuscript, and there is no single “correct” evidence span to retrieve. Here, accuracy naturally drops relative to \texttt{CMC}, since the system must both interpret the claim and find useful evidence without oracle guidance. Furthermore, unlike \texttt{CMC}, there is \emph{no defined oracle evidence span}, and thus retrieval recall cannot be reported. Two observations are particularly relevant:

\begin{itemize}[leftmargin=1.25em]
\item Overall accuracy is lower than on \texttt{CMC}, which is expected given the ambiguity in real reviewer text. This reflects the inherent difficulty of the task rather than a failure of the tool.
\item The dense retriever followed by a reranker provides the most reliable configuration in this setting (ACC~0.287 for \texttt{Qwen-2.5-7B}). Dense embeddings and reranking help bridge paraphrases and conceptual shifts that are common in reviewer phrasing, whereas purely sparse retrieval struggles when wording diverges from the manuscript.
\end{itemize}

The \texttt{CMC} and \texttt{RRC} results provide practical reassurance that the color-coded outputs shown in the interface of \tool\ correspond to a tested and empirically grounded pipeline. Users can therefore treat \tool\ not as a black box, but as a system whose behavior has been observed under both \textit{idealized} and \textit{realistic} review conditions.

\vspace{-1em}
\section{Concluding Remarks}
\label{sec:conclusion}

% As scientific conferences and journals continue to grow in size, the community faces increasing difficulty in ensuring that peer reviews remain grounded and fair. Reviewers, authors, and program committees all need support in determining whether evaluative statements in a review genuinely reflect what is written in the manuscript. \tool\ responds to this need by offering an automated and transparent framework that checks the factual grounding of reviews directly against the submitted paper.

% \tool\ will extract check-worthy claims from peer reviews, retrieve the most relevant passages from the manuscript, and verify each claim using an NLI-based method that identifies support, partial support, contradiction, or uncertainty. \tool\ accepts manuscript's PDFs as input as well as accepting papers from OpenReview. \tool\ inspect color-coded evidence highlights, and explore how each component of the pipeline behaves in practice. We also shared empirical findings that confirm the reliability of the underlying models of \tool, showing that the demo reflects a tested and validated system.
% We expect \tool\ to be helpful for reviewers who want to provide more evidence-based feedback, for authors preparing rebuttals or revisions, and for program committee members working to maintain consistent review quality at scale. By making claim verification accessible and interpretable, \tool\ aims to support a more accountable and transparent peer review process.

As conferences and journals continue to grow, it becomes increasingly difficult to ensure that peer reviews remain grounded and fair. Reviewers may misstate what a paper contains or overlook details, and these factual claims can influence decisions, rebuttals, and revisions. This creates a timely need for transparent methods that connect review statements to evidence in the submitted manuscript. \tool\ addresses this need by treating reviewer claims as document grounded retrieval and verification queries. The system extracts check worthy claims from reviews, retrieves the most relevant passages from the paper, and verifies them via an NLI style formulation. Results are presented in an interactive interface that highlights evidence directly in the manuscript, enabling rapid inspection of claim evidence alignment.

At the conference, we will demonstrate its end to end use on realistic review scenarios, including interactive exploration of evidence highlights and the behavior of the retrieval and verification pipeline. We will also showcase the modular design that allows alternative retrievers, rerankers, and verifiers to be substituted within the same framework. By providing an open, tested, and extensible system for claim level review verification, \tool\ offers a practical playground for the IR community to explore future research on retrieval and inference applied to scholarly communication.

%%
%% The next two lines define the bibliography style to be used, and
%% the bibliography file.

\bibliographystyle{ACM-Reference-Format}
\bibliography{sample-base}

\end{document}